# CALCIUM REMOVAL FROM CARDIAC CT IMAGES USING DEEP CONVOLUTIONAL NEURAL NETWORK

*Siming Yan[1,2], Feng Shi[1], Yuhua Chen[1,3], Damini Dey[1], Sang-Eun Lee[1,4], Hyuk-Jae Chang[4]*
*Debiao Li[1,3], Yibin Xie[1]*

[1]Biomedical Imaging Research Institute, Cedars Sinai Medical Center, Los Angeles, CA, USA
[2]Department of Computer Science, Peking University, Beijing, China
[3]Department of Bioengineering, University of California, Los Angeles, CA, USA
[4]Division of Cardiology, Yonsei University College of Medicine, Seoul, Korea

**ABSTRACT**

Coronary calcium causes beam hardening and blooming artifacts on cardiac computed tomography angiography (CTA) images, which lead to overestimation of lumen stenosis and reduction of diagnostic specificity. To properly remove coronary calcification and restore arterial lumen precisely, we propose a machine learning-based method with a multi-step inpainting process. We developed a new network configuration, Dense-Unet, to achieve optimal performance with low computational cost. Results after the calcium removal process were validated by comparing with gold-standard X-ray angiography. Our results demonstrated that removing coronary calcification from images with the proposed approach was feasible, and may potentially improve the diagnostic accuracy of CTA.

## 1. INTRODUCTION

Computed tomography angiography (CTA) allows high resolution depiction of blood vessels in vivo and is widely used in clinical practice. It is the current standard of care method for evaluating coronary artery atherosclerosis [1, 2]. CTA is a noninvasive procedure and much less costly and time-consuming than catheter-based invasive X-ray angiography. However, one of the major limitations of CTA is the poor diagnostic accuracy for segments with severe calcification. Because of high X-ray absorption, calcification leads to beam hardening artifacts and blooming artifacts on CTA, which affect the visualization of arterial lumen and make it difficult to accurately evaluate the degree of stenosis. Severe calcification often leads to significant overestimation of the disease severity, which further causes unnecessary catheterization and overtreatment.

A few studies have tried to address the calcification problem and they could be roughly categorized into two groups. The first group of studies focus on specialized image acquisitions. For example, Yoshioka et al. [3] proposed to acquire two specific scans with proper registration and generate the final CTA through subtraction to remove the calcification. Another group of methods focus on the segmentation of coronary calcium on CTA and retrospective removal. It does not require additional scans therefore does not involve additional radiation exposure or scan-related cost. However, these methods to remove irregular objects from the image may have limited performance in this scenario for several practical issues [4]. First, the boundaries of the calcified segments need to be manually segmented which may be impractical because of the large number. Second, partial volume effect hampers a precise definition of calcification boundaries, causing the segmentation either too small, leaving residual calcification, or too large resulting overestimation of the degree of lumen stenosis.

In this paper, we propose to approach the calcium removal process as an inpainting problem [5, 6]. Inpainting is the process of reconstructing the lost or deteriorated parts of an image. In essence, to restore a missing part in an image, inpainting methods hypothesize that the surrounding structures are continuous and thus the missing information can be extrapolated through the context.

Existing methods for inpainting problem include diffusion based, exemplar-based, hierarchical super-resolution- based and spatial patch blending approaches [6]. In this work, we propose to address the inpainting problem using the recently emerged convolutional neural network (CNN) [7]. CNN is comprised of many layers of learning units called neurons and these neurons could be effectively trained to extract the abstract features in an image, as a cost of millions to billions of parameters. Auto-encoders is a classic CNN-based method for image inpainting [8]. It consists of an encoder path capturing the context of an image into a compact latent feature representation and a decoder path where the representation is used to produce the missing image content. To optimize the performance and computational efficiency for calcium removal, we developed a new network architecture based on the aforementioned networks. The rest of the paper is organized as: in Section 2, our proposed approach is presented in detail; in Section 3, experimental results are reported; in Section 4, conclusions of our study are summarized.

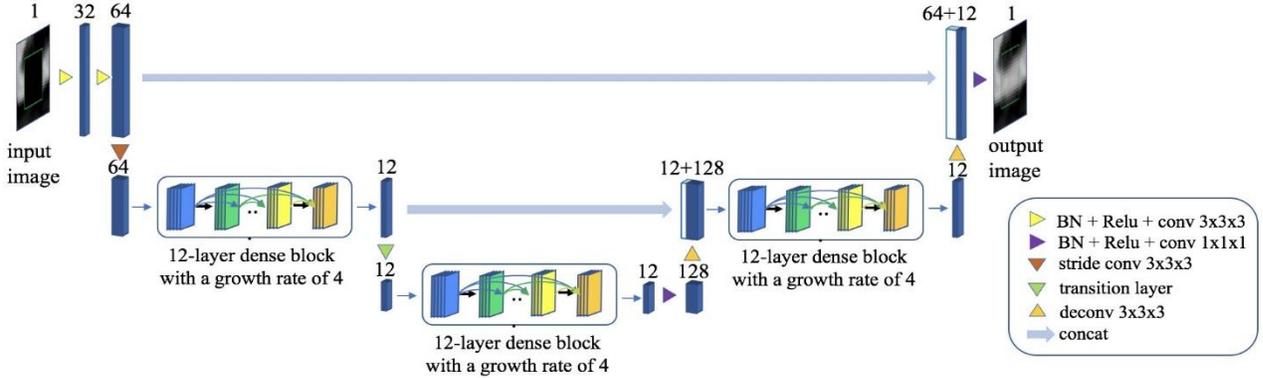

**Fig. 1.** The schematic of the proposed Dense-Unet. Three 12-layer dense blocks are shown in the network.

## 2. PROPOSED APPROACH

To formulate the coronary calcification removal process as an inpainting problem, we treated the regions with calcification as deteriorated parts of the image. Also, since the calcification boundary was irregular and difficult to draw, small regular cubic masks were used to remove and reconstruct the calcified segments, in a sliding window manner. The input CT images were divided into overlapping 3D patches and an in-painting mask was defined in the center of each patch. We used fixed size inpainting mask at the center of the image, so that information of surrounding structures could be consistently used to restore the missing information. For CNN architecture, we propose an end-to-end architecture, Dense-Unet, which uses a Unet-like structure [9, 10] with encoder and decoder paths and shortcut connections, as well as injected dense blocks [11] to significantly reduce the number of parameters and accelerate the training process. Details of our approach are provided in the following subsections.

### 2.1. Data Preparation

Our training dataset includes CT images from 60 patients with isotropic 0.47mm resolution, acquired by a clinical scanner (SOMATOM Definition Flash, Siemens, Germany). The volume of calcification was relatively small, as most of which are about 100 voxels. Small patches with the size of 32x32x32 were extracted from the original image. Since calcified segments were located in the center region of 160x160x64 in every image, patches were sampled in this area at a frequency of 10x10x4 in each dimension with 50% overlapping. Together, we have 24000 patches from all subjects. These patches were further flipped in each of the three dimensions for data augmentation. We then defined the inpainting mask as a cubic box in the center of each patch with size of 16x16x16. The size and position of the mask was fixed to simplify and reduce the difficulty of the image restoration process.

Calcification and bones had the highest intensity in CTA images and there was no bone in the center area as we defined above. Thus, we were able to locate the calcification region with a threshold of 700 Hounsfield Unit (HU). The patches without calcification were also used in the training process in order to allow the network to restore non-calcified segments. In the later testing stage, patches with calcium present were provided to the model with the calcium covered by the mask so the output image would also be calcium-free.

### 2.2. Proposed Dense-Unet

We propose a novel Dense-Unet architecture as shown in **Fig. 1.** The inputs were the patches with inpainting mask. Similar to the U-net, all images went through a contracting path that was composed of convolution layers and pooling layers for hierarchical extraction of high- and low-order features. These features then underwent an expanding path to produce an upsampled result that had the same dimension as the input image. To improve the performance as well as reduce computational cost, we adopted the idea of Denseblocks from Dense-Net [11]. Three 12-layer Denseblock with a growth rate of 4 were used. Between Denseblocks, transformation layer was composed by batch normalization (BN), ReLU activation, and a 3x3x3 convolutional layer. The proposed Dense-Unet was implemented in Tensorflow [12]. All the convolution layers were padded and the output image had the same size as the input image. The loss function of our network was mean squared error (MSE). The optimization was realized with Adam gradient descent optimization. The initial learning rate was $1e^{-4}$ with a batch size of 3. The growth rate of denseblock was 4. A total of 70k training steps were performed in 4 hrs

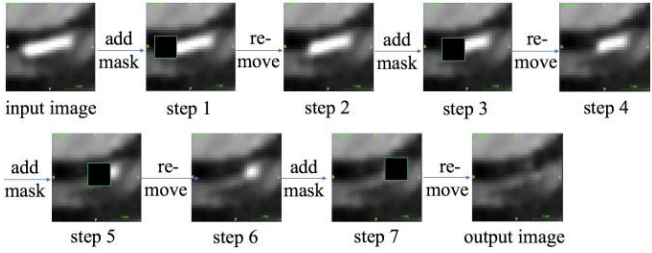

**Fig. 2.** Illustration of the proposed multi-step calcium removal process. A regional inpainting problem was solve progressively along the entire calcified arterial segment until all calcium is removed.

using a workstation equipped with Nvidia GeForce GTX 1080Ti 8GB GPU.

### 2.3. Sliding Window Strategy

Ideally, the inpainting mask should cover the entire calcified region to thoroughly remove all calcium. In practice, there were thin and long calcified segments exceeding the mask size. Therefore, we designed a sliding window strategy. For example, as shown in **Fig. 2**, the inpainting mask moved from left to right as a sliding window, and gradually restored the image. Results showed that it was effective for a long segment recovery. **Fig. 3** includes more examples of restored arterial lumen after calcium removal.

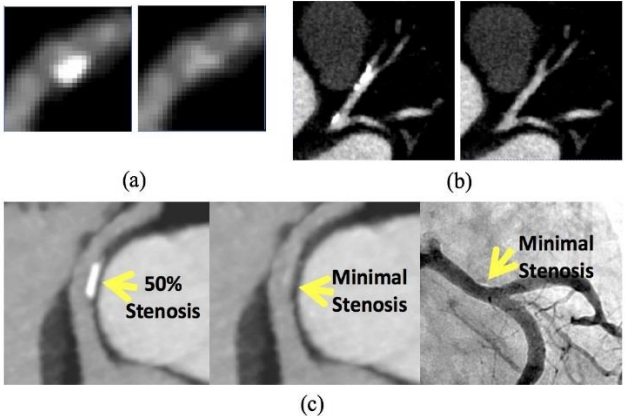

**Fig. 3.** Examples of restored arterial lumen after calcium removal. In (a)-(b), left figure shows the original lumen with the presence of calcification and right figure shows the restored lumen after removing calcification. (c) is a representative case in Experiment 2. The original CT showed 50% lumen stenosis at a calcified plaque on the left main segment. Calcium-removed CT showed minimal lumen stenosis at this location, which matched the finding from the X-ray angiography.

### 3. EXPERIMENTS AND RESULTS

Besides our training dataset of the images from 60 patients CTA images, we introduced additional images from 10 patients for testing. As there was no ground truth of calcium-removed CTA image as reference, we design two experiments to validate our test results. In the first experiment, we evaluated the restoration performance at non-calcified part in the testing images. The logic of this experiment is based on the assumption that if the unseen mask in testing images could be properly restored, our method is partially justified that it may work when the mask contains calcification. In the second experiment, we quantified the level of lumen stenosis in the original and calcium-removed images, respectively, and compared the results with the gold standard X-ray angiography (XA).

In Experiment 1, we randomly chose 20 patches with the size of 32x32x32 voxels and the inpainting mask at the center of patch with size of 16x16x16. The recovering accuracy was evaluated using MSE between the restored patch and the original patch. We compared the performance between the proposed Dense-Unet, Autoencoders, Densenet and U-net. Results of a sample patch is shown in **Fig. 4**, in which the results of Autoencoders showed a gap in the center of the recovered mask, whereas the proposed method provided the most consistent restoration. **Fig. 5(a)** shows the MSE per patch for all methods in comparison. The proposed network achieved the lowest MSE among all methods. We further tested the strategies of computing MSE for the entire output image or only the inpainting mask, and the result showed that the former was better (MSE per patch 407.55 vs 4077.78). This met our expectation that the surrounding of image could play a role as regularization to make the restoring region less blurry. In addition, we compared the settings of using size of 16 inpainting mask verses size of 8, and results showed that the latter was better with lower MSE (407.55 vs 25.03).
In Experiment 2, we randomly selected 7 patients with severely calcified coronary segments observed on CTA. Severe calcification was defined as more than one quadrant calcification on a cross-sectional image of the vessel on CTA. The degree of lumen stenosis of these segments was independently quantified on the original CTA images and calcium-removed CTA images, using a previously validated semi-automatic quantification software (AutoPlaq version 9.7, Cedars-Sinai Medical Center, USA) [13]. The degree of stenosis was then compared with the findings of the gold-standard quantitative invasive coronary X-ray angiography (XA). **Fig. 3(c)** shows a representative case for this experiment. The absolute differences in lumen degree stenosis measurements between the original CTA vs XA, as well as calcium-removed CTA vs XA, were calculated and plotted in **Fig. 5(b)**. On average CTA images had approximately 20% overestimation of stenosis degree whereas calcium-removed CT showed none.

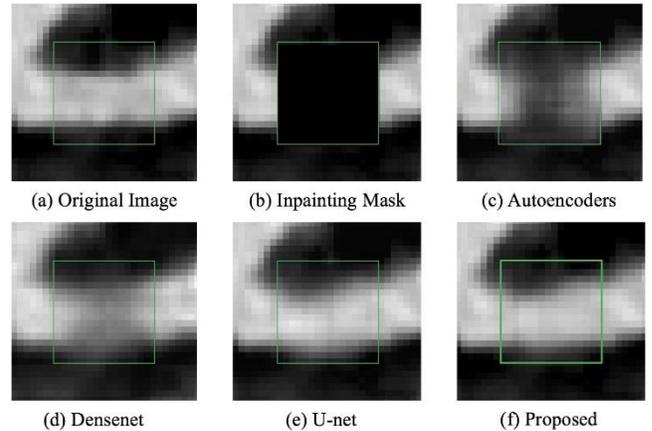

**Fig. 4**. Examples of the inpainting process using different approaches. The green box is the part to be recovered. (a) shows the original image; (b) shows the image with the inpainting mask as the input to the network; (c) uses Auto-encoders which shows a blurred result; (d) uses Densenet, which shows minor improvements; (e) uses Unet (f) uses our proposed Dense-Unet.

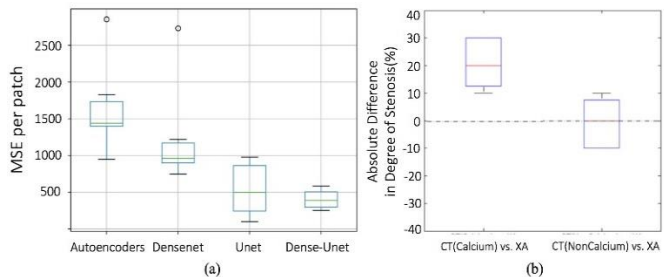

**Fig. 5.** Summary results of the experiments. (a) shows the boxplot comparing the performance of the four methods tested for 20 images. Vertical axis refers to the MSE values. The green line refers to the mean value of every methods and black line refers to the max and min value. The proposed Dense-Unet showed the lowest mean MSE. (b) shows the boxplot comparing the absolute differences in degree of stenosis measurements between original CT vs. X-ray Angiography (XA), and calcium-removed CT vs. XA, respectively, of 7 randomly selected calcified lesions. The original CT overestimated stenosis whereas calcium-removed CT did not.

## 4. LIMITATIONS AND FUTURE WORK

Despite our best efforts, the sizes of training and test dataset remain relatively small and homogenous in this study due to the availability of patient data to us. To reduce the potential risk of bias, a larger set of CTA images, ideally acquired at different sites and/or using different scanners, would be useful in the future to potentially improve the training and validate the results of our method in a more robust fashion. Experiment 2 in this study should be extended in the future to include more subjects with gold standard reference, which would allow formal evaluation of diagnostic accuracy such as sensitivity, specificity and receiver operating characteristic.

## 5. CONCLUSION

In this work, we demonstrated the feasibility of removing calcification in the cardiac CT images using deep neural network. The proposed new network configuration, Dense-Unet, achieved superior performance compared with existing network structures. The results of lumen restoration after removing calcification appeared promising and further validation against clinical reference is warranted.